\documentclass[10pt,twocolumn,letterpaper]{article}

\usepackage{cvpr}              
\usepackage{multirow}
\usepackage{booktabs}
\usepackage{pifont}
\usepackage{colortbl}
\usepackage{algorithm}
\usepackage{algorithmic}
\usepackage[export]{adjustbox}
\usepackage[accsupp]{axessibility}
\definecolor{my_green}{RGB}{51,102,0}
\definecolor{my_red}{RGB}{204, 0, 0}

\definecolor{paired-light-blue}{RGB}{198, 219, 239}
\definecolor{paired-dark-blue}{RGB}{49, 130, 188}
\definecolor{paired-light-orange}{RGB}{251, 208, 162}
\definecolor{paired-dark-orange}{RGB}{230, 85, 12}
\definecolor{paired-light-green}{RGB}{199, 233, 193}
\definecolor{paired-dark-green}{RGB}{49, 163, 83}
\definecolor{paired-light-purple}{RGB}{218, 218, 235}
\definecolor{paired-dark-purple}{RGB}{117, 107, 176}
\definecolor{paired-light-gray}{RGB}{217, 217, 217}
\definecolor{paired-dark-gray}{RGB}{99, 99, 99}
\definecolor{paired-light-pink}{RGB}{222, 158, 214}
\definecolor{paired-dark-pink}{RGB}{123, 65, 115}
\definecolor{paired-light-red}{RGB}{231, 150, 156}
\definecolor{paired-dark-red}{RGB}{131, 60, 56}
\definecolor{paired-light-yellow}{RGB}{231, 204, 149}
\definecolor{paired-dark-yellow}{RGB}{141, 109, 49}  
\definecolor{myblue}{RGB}{218,232,252}
\definecolor{mygray}{RGB}{220,220,220}
\definecolor{mypink}{RGB}{251,49,153}

\definecolor{cvprblue}{rgb}{0.21,0.49,0.74}
\usepackage[pagebackref,breaklinks,colorlinks,allcolors=cvprblue]{hyperref}

\def\paperID{1255} 
\def\confName{CVPR}
\def\confYear{2025}

\title{WF-VAE: Enhancing Video VAE by Wavelet-Driven Energy Flow for Latent Video Diffusion Model}

\author{
    Zongjian Li\textsuperscript{\rm 1,\rm 3,*},
    Bin Lin\textsuperscript{\rm 1,\rm 3,\thanks{Equal contribution}},
    Yang Ye\textsuperscript{\rm 1,\rm 3},
    Liuhan Chen\textsuperscript{\rm 1,\rm 3},
    Xinhua Cheng\textsuperscript{\rm 1,\rm 3}, \\
    Shenghai Yuan\textsuperscript{\rm 1,\rm 3},
    Li Yuan\textsuperscript{\rm 1,\rm 2,\thanks{Corresponding author.}} \\  \\
    \textsuperscript{1}Shenzhen Graduate School, Peking University, \textsuperscript{2}Peng Cheng Laboratory, \\ \textsuperscript{3}Rabbitpre Intelligence
}

\begin{document}

\maketitle


\begin{abstract}
Video Variational Autoencoder (VAE) encodes videos into a low-dimensional latent space, becoming a key component of most Latent Video Diffusion Models (LVDMs) to reduce model training costs. 
However, as the resolution and duration of generated videos increase, the encoding cost of Video VAEs becomes a limiting bottleneck in training LVDMs. Moreover, the block-wise inference method adopted by most LVDMs can lead to discontinuities of latent space when processing long-duration videos.
The key to addressing the computational bottleneck lies in decomposing videos into distinct components and efficiently encoding the critical information. Wavelet transform can decompose videos into multiple frequency-domain components and improve the efficiency significantly, we thus propose \textbf{Wavelet Flow VAE (WF-VAE)}, an autoencoder that leverages multi-level wavelet transform to facilitate low-frequency energy flow into latent representation. Furthermore, we introduce a method called \textbf{Causal Cache}, which maintains the integrity of latent space during block-wise inference.
Compared to state-of-the-art video VAEs, WF-VAE demonstrates superior performance in both \textbf{PSNR} and \textbf{LPIPS} metrics, achieving \textbf{2×} higher throughput and \textbf{4×} lower memory consumption while maintaining competitive reconstruction quality. 
Our code and models are available at \href{https://github.com/PKU-YuanGroup/WF-VAE}{https://github.com/PKU-YuanGroup/WF-VAE}.
\end{abstract}

\vspace{-1.2em}    
\section{Introduction}
\label{sec:intro}

The release of Sora~\cite{videoworldsimulators2024}, a video generation model developed by OpenAI, has pushed the boundaries of synthesizing photorealistic videos, drawing unprecedented attention to the field of video generation. Recent advancements in Latent Video Diffusion Models (LVDMs), such as Open-Sora Plan~\cite{pku_yuan_lab_and_tuzhan_ai_etc_2024_10948109}, Open-Sora~\cite{opensora}, CogVideoX~\cite{yang2024cogvideox}, EasyAnimate~\cite{xu2024easyanimate}, Movie Gen~\cite{polyak2024moviegencastmedia}, and ConsisID~\cite{yuan2024identity}, have led to substantial improvements in video generation quality. These methods establish a compressed latent space~\cite{rombach2022high} using a pre-trained video Variational Autoencoder (VAE), where the compression quality fundamentally determines the generative performance.

\begin{figure}[t]
  \centering  
  \vspace{1em}
  \includegraphics[width=0.93\linewidth]{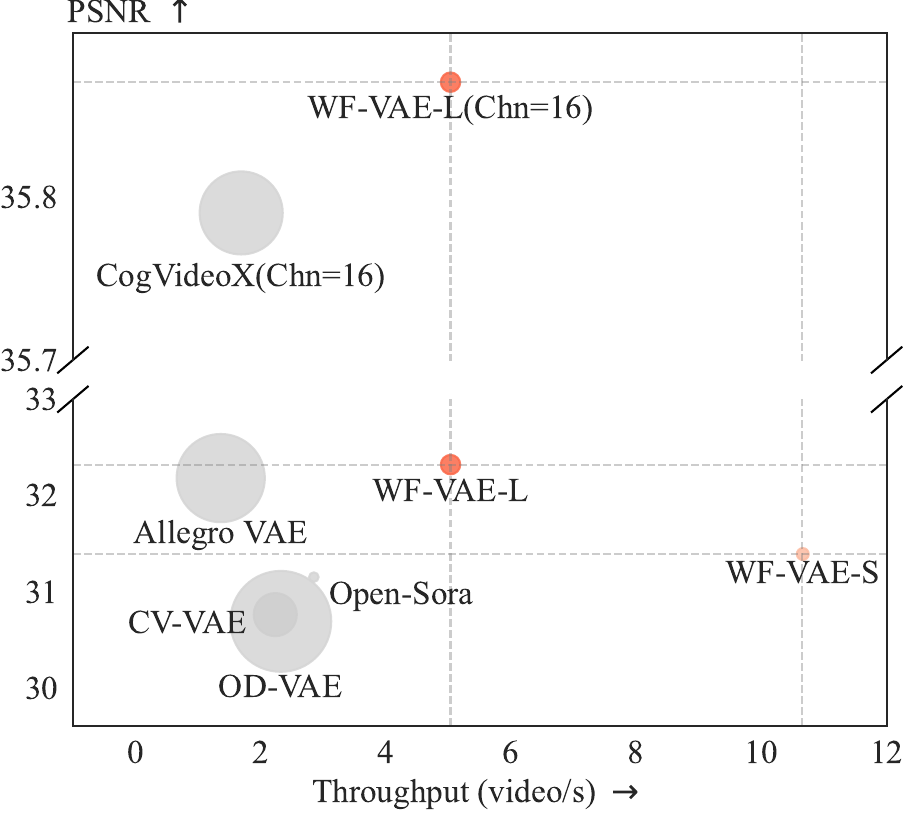} 
  \caption{\textbf{Performance comparison of video VAEs.} Bubble area indicates the memory usage during inference. All measurements are conducted on 33 frames with 256×256 resolution videos. ``Chn'' represents the number of latent channels. Higher PSNR and throughput indicate better performance. }  
  \label{fig:head_comparision}
  \vspace{-1.4em}
\end{figure}

Current video VAEs remain constrained by fully convolutional architectures inherited from the image era. They address video flickering and redundant information by incorporating spatio-temporal interaction and spatio-temporal compression layers. Several recent works, including OD-VAE~\cite{pku_yuan_lab_and_tuzhan_ai_etc_2024_10948109,chen2024od}, CogVideoX~\cite{yang2024cogvideox}, CV-VAE~\cite{zhao2024cv}, and Allegro~\cite{zhou2024allegroopenblackbox} adopt dense 3D structure to achieve high-quality video compression. While these methods demonstrate impressive reconstruction performance, they require prohibitively intensive computational resources. In contrast, alternative approaches such as Movie Gen~\cite{polyak2024moviegencastmedia} and Open-Sora~\cite{opensora} utilize 2+1D architecture, resulting in reduced computational requirements at the cost of lower reconstruction quality. Previous architectures have inadequately leveraged temporal redundancy in video data, necessitating the use of redundant spatio-temporal interaction layers to improve video compression quality.

Many attempts employ block-wise inference strategies to trade computation time for memory, thus addressing computational bottlenecks in processing high-resolution, long-duration videos. EasyAnimate~\cite{xu2024easyanimate} introduced Slice VAE to encode and decode video frames in groups, but this approach leads to discontinuous output videos. Several methods, including Open-Sora~\cite{opensora}, Open-Sora Plan~\cite{pku_yuan_lab_and_tuzhan_ai_etc_2024_10948109}, Allegro~\cite{zhou2024allegroopenblackbox}, implement tiling inference strategies but often produce spatio-temporal artifacts in overlapping regions. Although CogVideoX~\cite{yang2024cogvideox} employ caching to ensure convolution continuity, its reliance on group normalization~\cite{wu2018groupnormalization} disrupts the independence of temporal feature, thus preventing lossless block-wise inference. These limitations highlight two core challenges faced by video VAEs: (\textbf{1}) \textit{excessive computational demands due to redundant architecture}, and (\textbf{2}) \textit{compromised latent space integrity resulting from existing tiling inference strategies, which cause artifacts and flickering in reconstructed videos.}

Wavelet transform~\cite{Mallat_1989,prochazka2011three} decomposes videos into multiple frequency-domain components. This decomposition enables prioritization strategies for encoding crucial video components. In this work, we propose \textbf{Wavelet Flow VAE (WF-VAE)}, a novel autoencoder that utilizes multi-level wavelet transforms for extracting multi-scale pyramidal features and establishes a main energy flow pathway for these features to flow into latent representation. This pathway bypasses low-frequency video information to latent space, skipping the backbone network. Our WF-VAE enables a simplified backbone design with reduced 3D convolutions, significantly reducing computational costs. To address the potential latent space disruption, we propose \textit{Causal Cache} mechanism. This approach leverages the properties of causal convolution. It maintains the continuity of the convolution sliding window through a caching strategy, which ensures numerical identity between block-wise inference and direct inference results. Experimental results show that WF-VAE achieves state-of-the-art reconstruction quality and computational efficiency performance. To summarize, major contributions of our work include:
\begin{itemize}
\item We propose \textbf{WF-VAE}, which leverages multi-level wavelet transforms to extract pyramidal features and establishes a main energy flow pathway for video information flow into a latent representation.
\item We introduce a lossless block-wise inference mechanism called \textit{Causal Cache}, which maintains identical performance as direct inference across videos of any duration.
\item Extensive experimental evaluations of video reconstruction and generation demonstrate that WF-VAE achieves state-of-the-art performance.

\end{itemize}

\section{Related Work}
\label{sec:related_work}

\begin{figure*}[ht]
    \centering    \includegraphics[width=\linewidth]{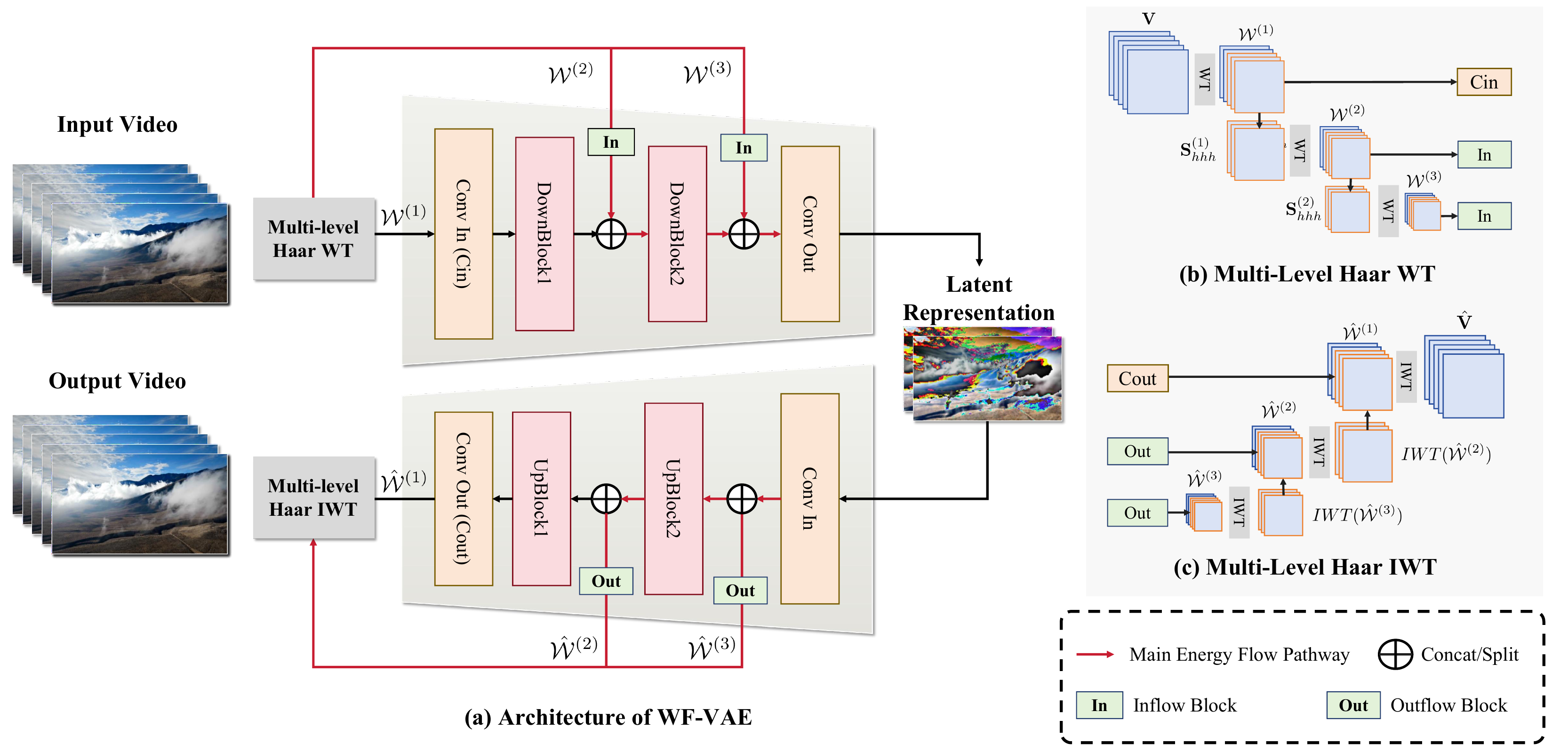}
    \vspace{-2.0em}
    \caption{\textbf{Overview of WF-VAE.} Our architecture consists of a backbone and a main energy flow pathway. The pathway functions as a “highway” for the main flow of video energy, channeling this energy into the backbone through concatenations, allowing more critical video information to be preserved in the latent representation.}
    \label{fig:Main}
    \vspace{-1.2em}
\end{figure*}

\noindent\textbf{Variational Autoencoders.}\quad
\cite{Kingma_Welling_2013} introduced the VAE based on variational inference, establishing a novel generative network structure. Subsequent research~\cite{Mittal_Engel_Hawthorne_Simon_2021, Vahdat_Kreis_Kautz_2021, Sinha_Song_Meng_Ermon_2021} demonstrated that training and inference in VAE latent space could substantially reduce computational costs for diffusion models. \cite{rombach2022high} further proposed a two-stage image synthesis approach by decoupling perceptual compression from diffusion model. After that, numerous studies explored video VAEs with a focus on more efficient video compression, including Open-Sora Plan~\cite{pku_yuan_lab_and_tuzhan_ai_etc_2024_10948109}, CogVideoX~\cite{yang2024cogvideox}, and other models~\cite{bao2024viduhighlyconsistentdynamic,yu2024languagemodelbeatsdiffusion,zhou2024allegroopenblackbox,zhao2024cv,chen2024od,zhu2023languagebind}. Current video VAE architectures primarily derive from earlier image VAE design~\cite{rombach2022high,esser2024scalingrectifiedflowtransformers} and use a convolutional backbone. LiteVAE~\cite{sadat2024litevae} employs a wavelet-based encoder inspired by latent-pixel similarity but overlooks energy perspectives, causing encoding-decoding energy asymmetry.

\noindent\textbf{Latent Video Diffusion Models.}\quad
In the early stages of Latent Video Diffusion Models (LVDMs) development, models like AnimateDiff~\cite{guo2023animatediff}, and MagicTime~\cite{yuan2024magictime, yuan2024chronomagic} primarily utilized the U-Net backbone~\cite{Ronneberger_Fischer_Brox_2015} for denoising, without temporal compression in VAEs. Following the paradigm introduced by Sora, recent open-source models such as Open-Sora \cite{opensora}, Open-Sora Plan \cite{pku_yuan_lab_and_tuzhan_ai_etc_2024_10948109,lin2023video}, and CogVideoX \cite{yang2024cogvideox} have adopted DiT backbone with a spatiotemporally compressed VAE. Some methods, including Open-Sora~\cite{opensora} and Movie Gen~\cite{polyak2024moviegencastmedia}, employ a 2+1D design in either the DiT backbone or the VAE to reduce training costs. For LVDMs, the upper limit of video generation quality is primarily determined by the VAE's reconstruction quality. As overhead increases rapidly with scale, optimizing the VAE becomes essential for processing large-scale data and enabling extensive pre-training.

\vspace{-0.1em}
\section{Method}

\subsection{Wavelet Transform}

\noindent\textbf{Preliminary.}\quad The Haar wavelet transform~\cite{Mallat_1989,prochazka2011three}, a fundamental form of wavelet transform, is widely used in signal processing~\cite{finder2024waveletconvolutionslargereceptive,gal2021swagan,huang2017wavelet, lhnet, lhnetv2}. It efficiently captures spatio-temporal information by decomposing signals through two complementary filters. The first is the Haar scaling filter $\mathbf{h} = \frac{1}{\sqrt{2}}[1, 1]$, which acts as a low-pass filter that captures the average or approximation coefficients. The second is the Haar wavelet filter $\mathbf{g} = \frac{1}{\sqrt{2}}[1, -1]$, which functions as a high-pass filter that extracts the detail coefficients. These orthogonal filters are designed to be simple yet effective, with the scaling filter smoothing the signal and the wavelet filter detecting local changes or discontinuities.

\noindent\textbf{Multi-level Wavelet Transform.}\quad For a video signal $\mathbf{V} \in \mathbb{R}^{c \times t \times h \times w}$, where $c$, $t$, $h$, and $w$ denote the number of channels, temporal frames, height, and width respectively, the 3D Haar wavelet transform at layer $l$ is defined as:

\begin{equation}
\mathbf{S}^{(l)}_{ijk} = \mathbf{S}^{(l-1)} * (f_i \otimes f_j \otimes f_k),
\end{equation}
where $f_i, f_j, f_k \in \{\mathbf{h}, \mathbf{g}\}$ represent the filters applied along each dimension, and $*$ represents the convolution operation. The transform begins with $\mathbf{S}^{(0)} = \mathbf{V}$, and for subsequent layers, $\mathbf{S}^{(l)} = \mathbf{S}_{hhh}^{(l-1)}$, indicating that each layer operates on the low-frequency component from the previous layer. At each decomposition layer $l$, the transform produces eight sub-band components: $\mathcal{W}^{(l)} = \{ \mathbf{S}_{hhh}^{(l)}, \mathbf{S}_{hhg}^{(l)}, \mathbf{S}_{hgh}^{(l)}, \mathbf{S}_{ghh}^{(l)}, \mathbf{S}_{hgg}^{(l)}, \mathbf{S}_{ggh}^{(l)}, \mathbf{S}_{ghg}^{(l)}, \mathbf{S}_{ggg}^{(l)} \}$. Here, $\mathbf{S}_{hhh}^{(l)}$ represents the low-frequency component across all dimensions, while $\mathbf{S}_{ggg}^{(l)}$ captures high-frequency details. To implement different downsampling rates in the temporal and spatial dimensions, a combination of 2D and 3D wavelet transforms can be implemented. Specifically, to obtain a compression rate of 4×8×8 (temporal×height×width), we can employ a combination of a two-layer 3D wavelet transform followed by a one-layer 2D wavelet transform.

\subsection{Architecture Design of WF-VAE}

Through analyzing different sub-bands, we find that video energy is mainly concentrated in the low-frequency sub-band $\mathbf{S}^{(1)}_{hhh}$. Based on this observation, we establish an energy flow pathway, as illustrated in~\cref{fig:Main}, so that low-frequency information can smoothly flow from video to latent representation during the encoding process and then flow back to the video during the decoding process. This would inherently allow the model to pay more attention to low-frequency information and apply higher compression rates to high-frequency details. With the additional path, we can reduce the computational cost brought by the dense 3D convolutions in the backbone.

Specifically, given a video $\mathbf{V}$, we apply multi-level wavelet transform to obtain pyramid features $\mathcal{W}^{(1)},\mathcal{W}^{(2)}$ and $\mathcal{W}^{(3)}$. We utilize $\mathcal{W}^{(1)}$ as the input for the encoder and the target output for the decoder. The backbone and multi-level wavelet transform downsample the feature maps simultaneously at every downsampling layer, enabling the concatenation of features from two branches in the backbone. We employ Inflow Block to transform the channel numbers of $\mathcal{W}^{(2)}$ and $\mathcal{W}^{(3)}$ to $C_{flow}$, which are then concatenated with feature maps from backbone. We compare $C_{flow}$ to the width of the energy flow pathway, as analyzed in~\cref{section:model_analysis}. On the decoder side, we maintain a structure symmetrical to the encoder. We split feature maps with $C_{flow}$ channels from the backbone and process them through Outflow Block to obtain ${\hat{\mathcal{W}}^{(2)}, \hat{\mathcal{W}}^{(3)}}$. To allow information to flow from the lower layer to the $hhh$ sub-band of the next layer, we have:

\begin{equation}
   \hat{\mathbf{S}}^{(2)}_{hhh} = {IWT}(\hat{\mathcal{W}}^{(3)}) + \hat{\mathbf{S}}^{(2)}_{outflow,hhh}.
\end{equation}

Similarly, at the decoder output layer:

\begin{equation}
    \hat{\mathbf{S}}^{(1)}_{hhh} = {IWT}(\hat{\mathcal{W}}^{(2)}) + \hat{\mathbf{S}}^{(1)}_{outflow,hhh}.
\end{equation}

Overall, we strategically design shortcuts to prioritize low-frequency information, thereby enhancing its representation within the latent space.

\subsection{Causal Cache}

\begin{figure}[h]
    \centering
    \begin{subfigure}[b]{\linewidth}
        \centering
        \includegraphics[width=\linewidth]{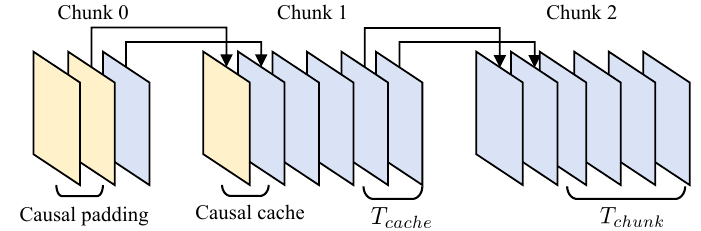}
        \vspace{-1.5em}
        \caption{Illustration of \textit{Casual Cache}.}
        \label{fig:ana-a}
    \end{subfigure}
    
    \begin{subfigure}[b]{\linewidth}
        \centering
        \includegraphics[width=\linewidth]{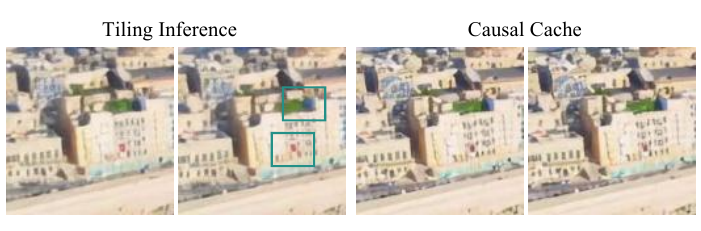}
                \vspace{-1.5em}
        \caption{Qualitative comparison of tiling inference and \textit{Causal Cache}}
        \label{fig:ana-b}
    \end{subfigure}
    \vspace{-1.8em}
    \caption{\textbf{(a)} \textit{Causal Cache} with a temporal kernel size of 3 and stride 1. \textbf{(b)} Comparison of tiling inference and \textit{Causal Cache}, highlighting how tiling causes locally \textbf{color and shape distortions} at overlaps, leading to global flickering in reconstructed videos.}
\end{figure}

We replace regular 3D convolutions with causal 3D convolutions~\cite{yu2024languagemodelbeatsdiffusion} in WF-VAE. The causal convolution applies $k_t - 1$ temporal padding at the start with kernel size $k_t$. This padding strategy ensures that the first frame remains independent from subsequent frames, thus enabling the processing of images and videos within a unified architecture. Furthermore, we leverage the causal convolution properties to achieve lossless inference. We first extract the initial frame from a video with $T$ frames. The remaining $T-1$ frames are then partitioned into temporal chunks, where $T_{chunk}$ represents the chunk size. Let $s_t$ denote the temporal convolutional stride, and $m = 0,1,2,\cdots$ represents the chunk block index. To maintain the continuity of convolution sliding windows, each chunk caches its tail frames for the next chunk. The number of cached frames is given by:
\begin{equation}
    T_{cache}(m) = k_t + mT_{chunk} - s_t \lfloor \frac{mT_{chunk}}{s_t} + 1 \rfloor.
\end{equation}

For example, $k_t = 3, s_t = 1, T_{chunk}=4$, the equation yields $T_{cache}(m) = 2$, as illustrated in~\cref{fig:ana-a}. Similarly, with $k_t = 3, s_t = 2, T_{chunk}=4$, we obtain $T_{cache}(m) = 1$, indicating only the last frame requires to be cached. Special cases exist, such as when $k_t=4, s_t=3, T_{chunk}=4$, which results in $T_{cache}(m)= (m \bmod 3 + 1)$. \cref{fig:ana-b} provides a qualitative comparison between \textit{Causal Cache} and the tiling strategy, illustrating how it effectively mitigates significant distortions in both color and shape.

\subsection{Training Objective}
\label{sec:training_objective}

Following the training strategies of \cite{rombach2022high,Esser_2021_CVPR}, our loss function combines multiple components, including reconstruction loss (comprising L1 and perceptual loss~\cite{Zhang_Isola_Efros_Shechtman_Wang_2018}), adversarial loss, and KL regularization \cite{Kingma_Welling_2013}. Our model is characterized by a low-frequency energy flow and symmetry between the encoder and decoder. To maintain this architectural principle, we introduce a regularization term denoted as $\mathcal{L}_{W\!L}$ (WL loss), which enforces structural consistency by penalizing deviations from the intended energy flow:
\begin{equation}
\mathcal{L}_{W\!L} = |\hat{\mathcal{W}}^{(2)} - \mathcal{W}^{(2)}| + |\hat{\mathcal{W}}^{(3)} - \mathcal{W}^{(3)}|.
\end{equation}

The final loss function is formulated as follows:
\begin{equation}
\begin{split}
    \mathcal{L} &= \mathcal{L}_{recon} +  \lambda_{adv} \mathcal{L}_{adv} + \lambda_{K\!L}\mathcal{L}_{K\!L} + \lambda_{W\!L} \mathcal{L}_{W\!L}.
\end{split}
\end{equation}

We examine the impact of the weighting factor $\lambda_{wl}$ in~\cref{section:model_analysis}. Following \cite{Esser_2021_CVPR}, we implement dynamic adversarial loss weighting to balance the relative gradient magnitudes between adversarial and reconstruction losses:
\begin{equation}
   \lambda_{\mathrm{adv}}=\frac{1}{2}\bigg(\frac{\|\nabla_{G_L}[\mathcal{L}_{\mathrm{recon}}]\|}{\|\nabla_{G_L}[\mathcal{L}_{\mathrm{adv}}]\|+\delta}\bigg),
\end{equation}
where $\nabla_{G_L}[\cdot]$ denotes the gradient with respect to last layer of decoder, and $\delta=10^{-6}$ is used for numerical stability.

\section{Experiments}

\subsection{Experimental Setup}

\noindent\textbf{Baseline Models.}\quad To assess the effectiveness of WF-VAE, we perform a comprehensive evaluation, comparing its performance and efficiency against several state-of-the-art VAE models. The models considered are: (1) OD-VAE~\cite{chen2024od}, a 3D causal convolutional VAE used in Open-Sora Plan 1.2~\cite{pku_yuan_lab_and_tuzhan_ai_etc_2024_10948109}; (2) Open-Sora VAE~\cite{opensora}; (3) CV-VAE~\cite{zhao2024cv}; (4) CogVideoX VAE~\cite{yang2024cogvideox}; (5) Allegro VAE~\cite{zhou2024allegroopenblackbox}; (6) SVD-VAE~\cite{blattmann2023stablevideodiffusionscaling}, which does not compress temporally and (7) SD-VAE~\cite{rombach2022high}, a widely used image VAE. Among these, CogVideoX VAE adopts a latent dimension of 16, whereas others utilize a latent dimension of 4. Notably, most VAEs have been validated on LVDMs, making them highly representative for comparison.

 \begin{table*}[ht]
 \centering
 \renewcommand{\arraystretch}{1.25}
 \scalebox{0.78}{
\begin{tabular}{c| cc cccc cccc}
 \toprule[1.2pt]
   \multirow{2}{*}{\textbf{Method}}  & \multirow{2}{*}{\textbf{TCPR}} & \multirow{2}{*}{\textbf{Chn}} & \multicolumn{4}{c}{\textbf{WebVid-10M}} & \multicolumn{4}{c}{\textbf{Panda-70M}}\\
   \cmidrule(lr){4-7} \cmidrule(lr){8-11}
   &&&  \textbf{PSNR} $(\uparrow)$ & \textbf{SSIM}$(\uparrow)$ & \textbf{LPIPS} $(\downarrow)$ & \textbf{rFVD} $(\downarrow)$ & \textbf{PSNR} $(\uparrow)$ & \textbf{SSIM} $(\uparrow)$ & \textbf{LPIPS}$(\downarrow)$ & \textbf{rFVD} $(\downarrow)$\\
\midrule
SD-VAE~\cite{rombach2022high}&$64(1\times8\times8)$&4&30.19&0.8377&0.0568&284.90 &30.46&0.8896&0.0395&182.99\\
SVD-VAE~\cite{blattmann2023stablevideodiffusionscaling}&$64(1\times8\times8)$&4&31.18&0.8689&0.0546&188.74&31.04&0.9059&0.0379&137.67\\
\midrule
CV-VAE~\cite{zhao2024cv}&$256(4\times8\times8)$&4&30.76&0.8566&0.0803&369.23 &30.18&0.8796&0.0672&296.28\\
OD-VAE~\cite{chen2024od}&$256(4\times8\times8)$&4&30.69&0.8635&0.0553&255.92&30.31&0.8935&0.0439&191.23\\
Open-Sora VAE ~\cite{opensora}&$256(4\times8\times8)$&4&31.14&0.8572&0.1001&475.23&31.37&0.8973&0.0662&298.47\\
Allegro ~\cite{zhou2024allegroopenblackbox}&$256(4\times8\times8)$&4&\underline{32.18}&\textbf{0.8963}&0.0524&209.68&\underline{31.70}&\textbf{0.9158}&0.0421&172.72\\
\rowcolor{cyan!3}
WF-VAE-S (Ours)&$256(4\times8\times8)$&4&31.39&0.8737&\underline{0.0517}&\underline{188.04}&31.27&0.9025&\underline{0.0420}&\underline{146.91}\\
\rowcolor{cyan!3}
WF-VAE-L (Ours)&$256(4\times8\times8)$&4&\textbf{32.32}&\underline{0.8920}&\textbf{0.0513}&\textbf{186.00}&\textbf{32.10}&\underline{0.9142}&\textbf{0.0411}&\textbf{146.24}\\
\midrule
CogVideoX-VAE~\cite{yang2024cogvideox}&$256(4\times8\times8)$&16&\underline{35.72}&\textbf{0.9434}&\underline{0.0277}&\underline{59.83}&\underline{35.79}&\underline{0.9527}&\underline{0.0198}&\underline{43.23}\\
\rowcolor{cyan!3}
WF-VAE-L (Ours)&$256(4\times8\times8)$&16&\textbf{35.76}&\underline{0.9430}&\textbf{0.0230}&\textbf{54.36}&\textbf{35.87}&\textbf{0.9538}&\textbf{0.0175}&\textbf{39.40} \\
\bottomrule[1.2pt]
\end{tabular}}
    \vspace{-0.4em}
 \caption{\textbf{Quantitative metrics of reconstruction performance.} Results demostrate that WF-VAE achieves state-of-the-art on reconstrcution performance comparing with other VAEs on WebVid-10M~\cite{Bain_Nagrani_Varol_Zisserman_2021} and Panda70M~\cite{Chen_2024_CVPR} datasets. TCPR represents the token compression rate, and Chn indicates the number of latent channels. The highest result is highlighted in \textbf{bold}, and the second highest result is \underline{underlined}.}
\label{tb:MainTable} 
\end {table*}

\begin{figure*}[ht]
    \centering    \includegraphics[width=\linewidth]{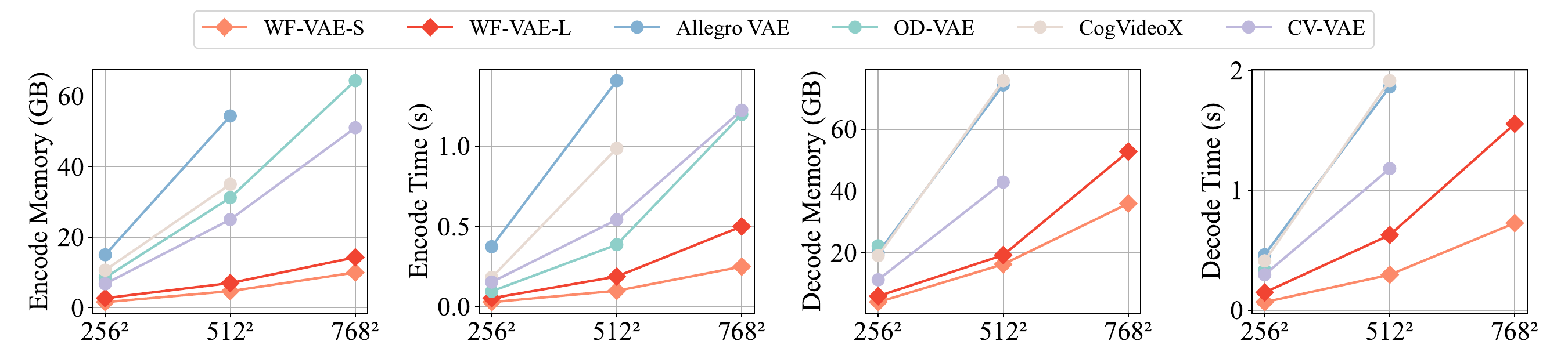}
    \vspace{-1.85em}
    \caption{\textbf{Computational performance of encoding and decoding}. We evaluate the encoding, decoding time, and memory consumption across 33 frames with 256×256, 512×512, and 768×768 resolutions (benchmark models without causal convolution are tested with 32 frames). WF-VAE surpasses other VAE models by a large margin in terms of both inference speed and memory efficiency.}
    \label{fig:SpeedMemoryTest}
    \vspace{-1.2em}
\end{figure*}

\noindent\textbf{Dataset \& Evaluation.}\quad
We utilize the Kinetics-400 dataset~\cite{kay2017kinetics} for both training and validation. For testing, we employ the Panda70M~\cite{Chen_2024_CVPR} and WebVid-10M~\cite{Bain_Nagrani_Varol_Zisserman_2021} datasets. To comprehensively evaluate the model's reconstruction performance, we select Peak Signal-to-Noise Ratio (PSNR)~\cite{Hore_Ziou_2010}, Learned Perceptual Image Patch Similarity (LPIPS)~\cite{Zhang_Isola_Efros_Shechtman_Wang_2018}, and Structural Similarity Index Measure (SSIM)~\cite{wang2004image} as primary evaluation metrics. Additionally, we use reconstruction Fréchet Video Distance (rFVD)~\cite{Unterthiner_Steenkiste_Kurach_Marinier_Michalski_Gelly_2019} to assess visual quality and temporal coherence. To assess our model's performance in generating results with the diffusion model, we utilize the UCF-101~\cite{Soomro_Zamir_Shah_2012} and SkyTimelapse~\cite{Xiong_Luo_Ma_Liu_Luo_2017} datasets for conditional and unconditional training 100,000 steps. Following~\cite{ma2024latte, skorokhodov2022stylegan}, we extract 16-frame clips of 2,048 videos to compute FVD$_{16}$. Additionally, we evaluate the Inception Score (IS)~\cite{Saito_Matsumoto_Saito_2017} exclusively on the UCF-101 dataset, as suggested by~\cite{ma2024latte}. 
We select Latte-L~\cite{ma2024latte} as the denoiser. Since we focus not on the generative performance but on whether the latent spaces of various video VAEs facilitate practical diffusion model training, we chose not to use the higher-performing Latte-XL.

\noindent\textbf{Training Strategy.}\quad
We employ the AdamW~\cite{Kingma_Ba_2014,loshchilov2019decoupledweightdecayregularization} optimizer with parameters $\beta_1 = 0.9$ and $\beta_2 = 0.999$, and set a fixed learning rate of $1 \times 10^{-5}$. Our training process comprises three stages: \textbf{(I)} the first stage aligns with~\cite{chen2024od}, where we preprocess videos to 25 frames with $256 \times 256$ resolution, and a total batch size of 8. \textbf{(II)} we refresh the discriminator, increase the number of frames to 49, and reduce the FPS by half to enhance motion dynamics. \textbf{(III)} we observe that a large $\lambda_{lpips}$ significantly affects video stability; therefore, we refresh the discriminator once more and set $\lambda_{lpips}$ to $0.1$. All three stages employ L1 loss: the initial stage is trained for 800,000 steps, while the subsequent stages are each trained for 200,000 steps. The training process utilizes 8 NVIDIA H100 GPUs. We implement a 3D discriminator and initiate GAN training from the start. All training hyperparameters are detailed in the appendix.

\begin{figure*}[ht]
    \centering    \includegraphics[width=0.96\linewidth]{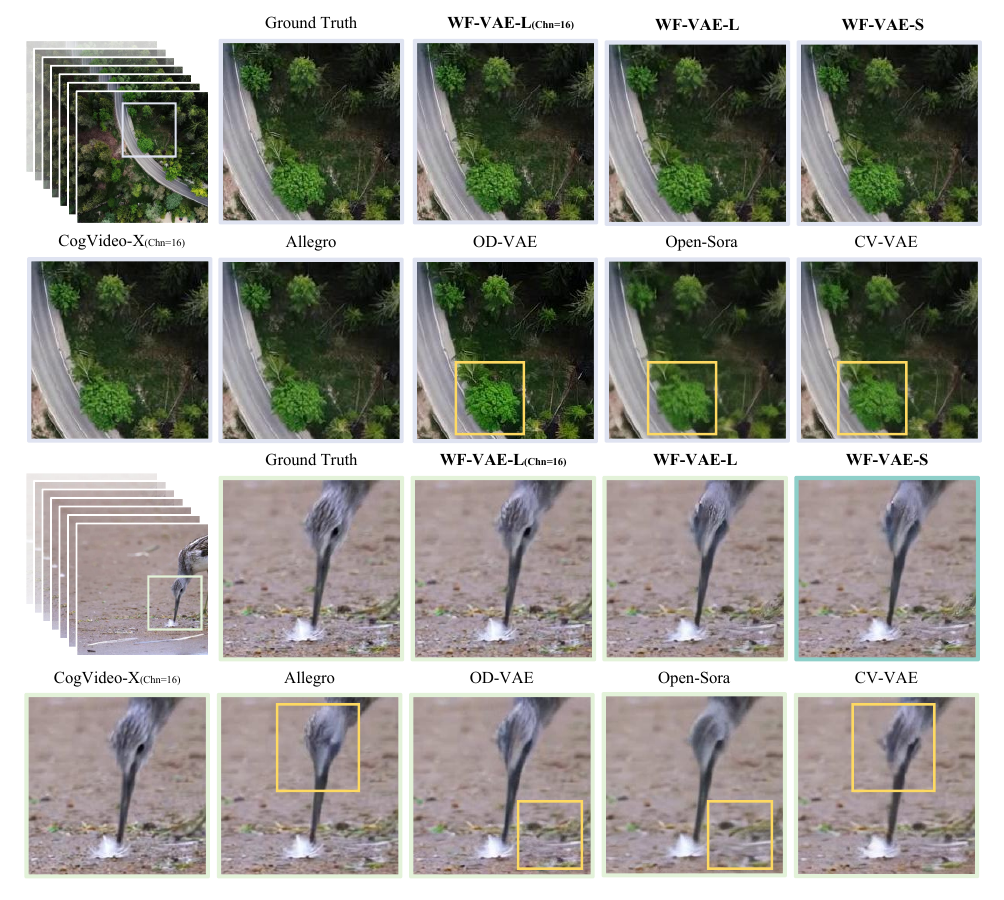}
    \vspace{-1.5em}

    \caption{\textbf{Qualitative comparison of reconstruction performance.} We select two scenarios to comprehensively evaluate the visual quality of videos reconstructed by existing VAEs. Top: scenario contains \textbf{rich details}. Bottom: scenario contains \textbf{fast motion}.}
    \label{fig:Recon}
    \vspace{-1.2em}
\end{figure*}

\subsection{Comparison With Baseline Models}
\label{sec:comparision}
We compare WF-VAE with baseline models in three key aspects: computational efficiency, reconstruction performance, and diffusion-based generation performance. To ensure fairness in comparing metrics and model efficiency, we disable block-wise inference strategies across all VAEs.

\noindent\textbf{Computational Efficiency.}\quad The computational efficiency evaluations are conducted using an H100 GPU with float32 precision. Performance evaluations are performed at 33 frames across multiple input resolutions. Due to non-causal convolution architecture, Allegro VAE is evaluated using 32-frame videos to maintain consistency. All benchmark VAEs use direct inference without block-wise inference strategies for a fair comparison. As shown in~\cref{fig:SpeedMemoryTest}, WF-VAE demonstrates superior inference performance compared to other VAEs. For instance, WF-VAE-L requires 7170 MB of memory for encoding a video at 512×512 resolution, whereas OD-VAE demands approximately 31944 MB (445\% higher). Similarly, CogVideoX consumes around 35849.33 MB (499\% higher), and Allegro VAE requires 55664 MB (776\% higher). These results highlight WF-VAE's significant advantages in large-scale training and data processing. In terms of encoding speed, WF-VAE-L achieves an encoding time of 0.0513 seconds, while OD-VAE, CogVideoX, and Allegro VAE exhibit encoding times of 0.0945 seconds, 0.1810 seconds, and 0.3731 seconds, respectively, which are approximately 184\%, 352\%, and 727\% slower. As illustrated in~\cref{fig:SpeedMemoryTest}, WF-VAE demonstrates notable advantages in computational efficiency during the decoding process.
 
\begin{figure*}[ht]  
    \centering  
    \begin{minipage}{0.48\textwidth}
        \centering
        \renewcommand{\arraystretch}{1.25}
        \scalebox{0.86}{
        \begin{tabular}{l|cccc}
            \toprule[1.2pt]
            \multirow{2}{*}{\textbf{Method}} & \multirow{2}{*}{\textbf{Chn}} & \multicolumn{1}{c}{\textbf{SkyTimelapse}} & \multicolumn{2}{c}{\textbf{UCF101}} \\
             \cmidrule(lr){3-3}\cmidrule(lr){4-5}
             & & \textbf{FVD}$_{16}$$\downarrow$ & \textbf{FVD}$_{16}$$\downarrow$  & \textbf{IS}$\uparrow$ \\
            \midrule
            Allegro~\cite{zhou2024allegroopenblackbox} & 4 & 117.28 & 1045.66 &  \underline{67.16} \\
            OD-VAE~\cite{chen2024od} & 4 & 130.79 & 1109.87 & 58.48 \\
            \rowcolor{cyan!3}
            WF-VAE-S (Ours) & 4 & \textbf{103.44} & \underline{1005.10} & 65.89 \\
            \rowcolor{cyan!3}
            WF-VAE-L (Ours) & 4 & \underline{113.67} & \textbf{929.55} & \textbf{70.53} \\
            \midrule
            CogVideoX~\cite{yang2024cogvideox} & 16 & 109.20 & 1117.57 & 57.47 \\
            \rowcolor{cyan!3}
            WF-VAE-L (Ours) & 16 & \textbf{108.69} & \textbf{947.18} & \textbf{71.86} \\
            \bottomrule[1.2pt]
        \end{tabular}
        }
        \vspace{-0.7em}
        \captionof{table}{\textbf{Quantitative evaluation of different VAE models for video generation.} We assess video generation quality using FVD$_{16}$ on both SkyTimelapse and UCF-101 datasets, and IS on UCF-101 following prior work \cite{ma2024latte}.}
        \label{tb:FVDResults}
    \end{minipage}%
    \hfill
    \begin{minipage}{0.48\textwidth}
        \centering
        \includegraphics[width=\linewidth]{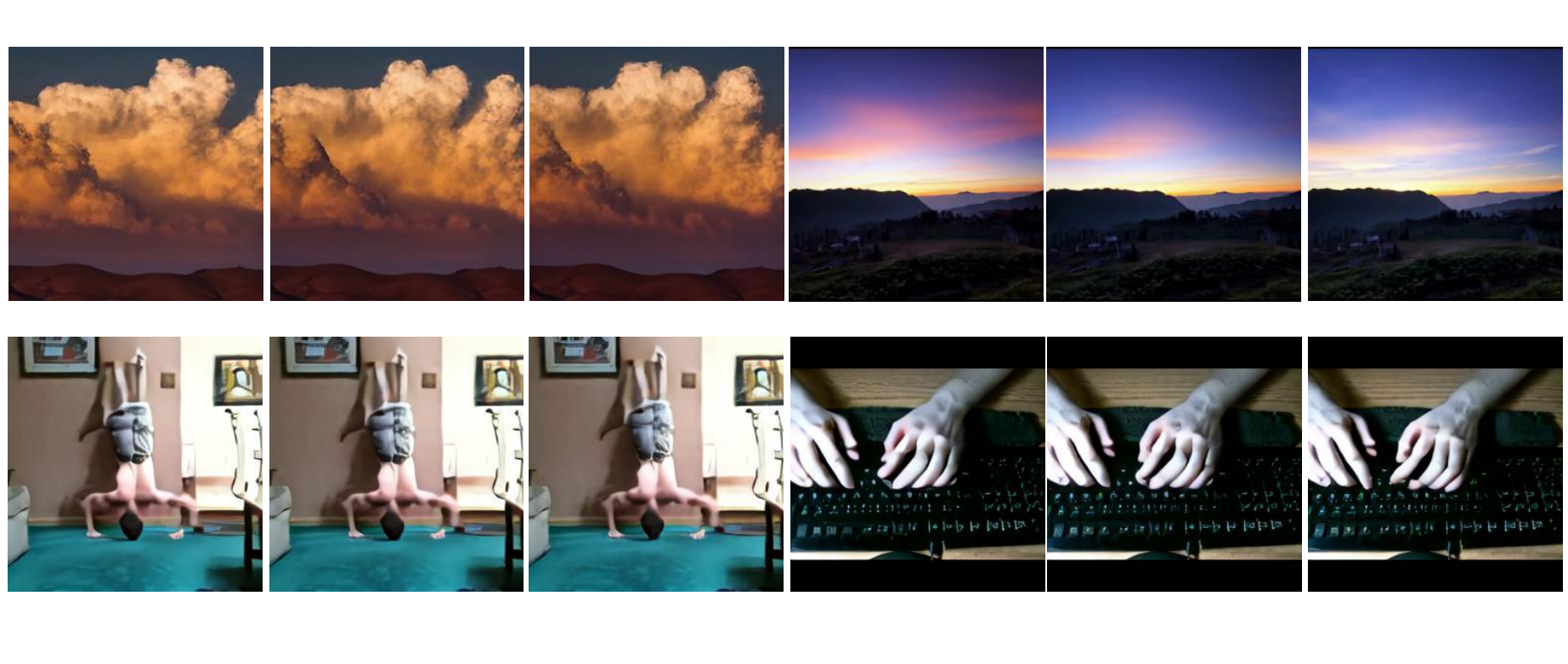}
        \vspace{-1.1em}

        \caption{\textbf{Generated videos using WF-VAE with Latte-L.} \textbf{Top:} results trained with the SkyTimelapse dataset. \textbf{Bottom:} results trained with the UCF-101 dataset.}
        \label{fig:LatteGenerate}
    \end{minipage}
    \vspace{-0.8em}
\end{figure*}

\begin{figure*}[ht]
    \centering
    \begin{subfigure}{0.32\textwidth}
        \centering
        \includegraphics[width=\textwidth]{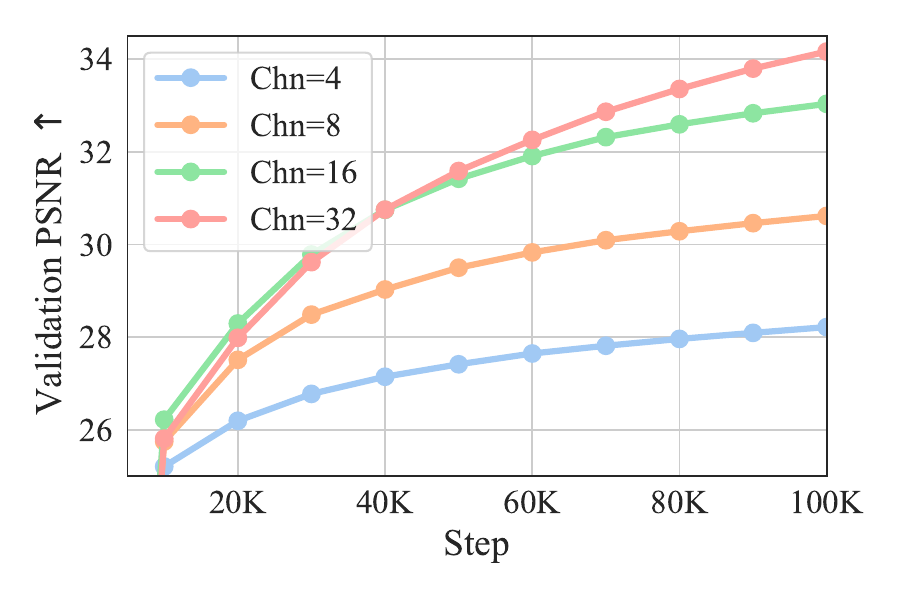}
        \vspace{-1.8em}
    \end{subfigure}
    \hfill
    \begin{subfigure}{0.32\textwidth}
        \centering
        \includegraphics[width=\textwidth]{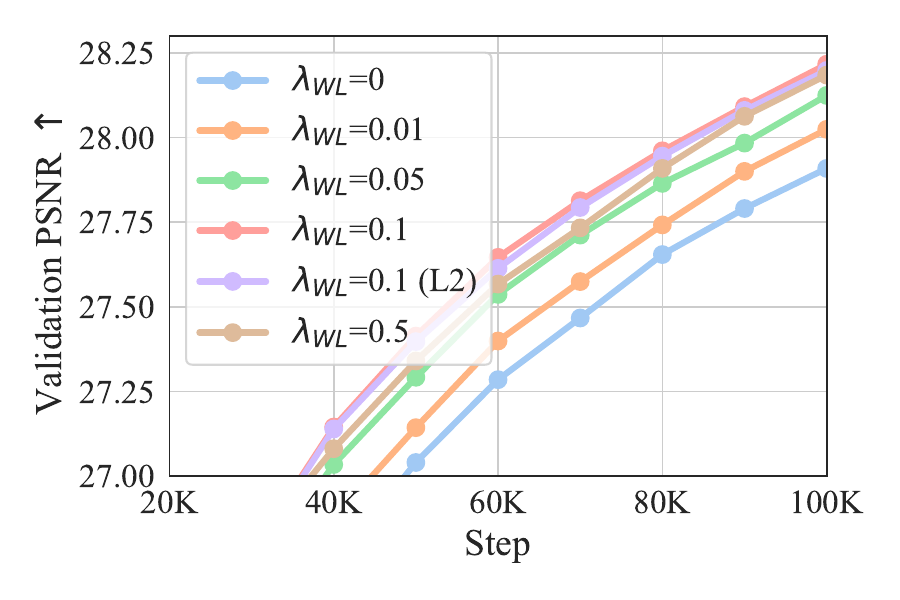}
        \vspace{-1.8em}
    \end{subfigure}
    \hfill
    \begin{subfigure}{0.32\textwidth}
        \centering
        \includegraphics[width=\textwidth]{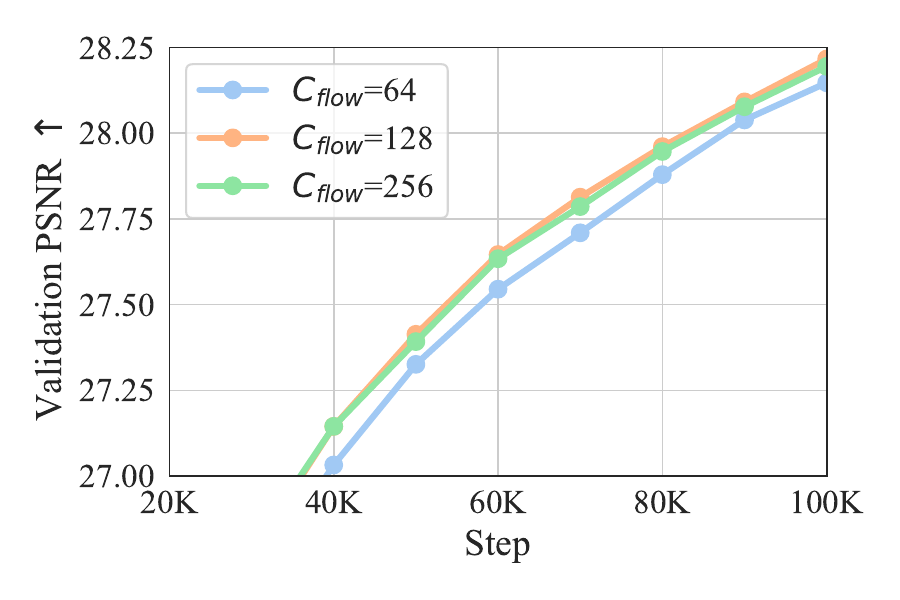}
        \vspace{-1.8em}
    \end{subfigure}
    
    \begin{subfigure}{0.32\textwidth}
        \centering
        \includegraphics[width=\textwidth]{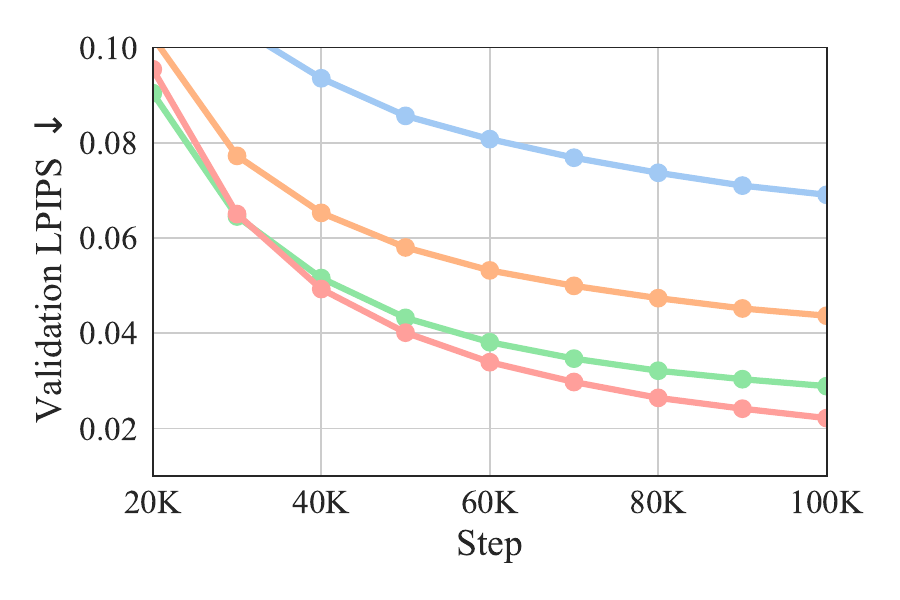}
        \vspace{-1.8em}
        \caption{Number of latent channels.}
        \label{fig:analysis-a}
    \end{subfigure}
    \hfill
    \begin{subfigure}{0.32\textwidth}
        \centering
        \includegraphics[width=\textwidth]{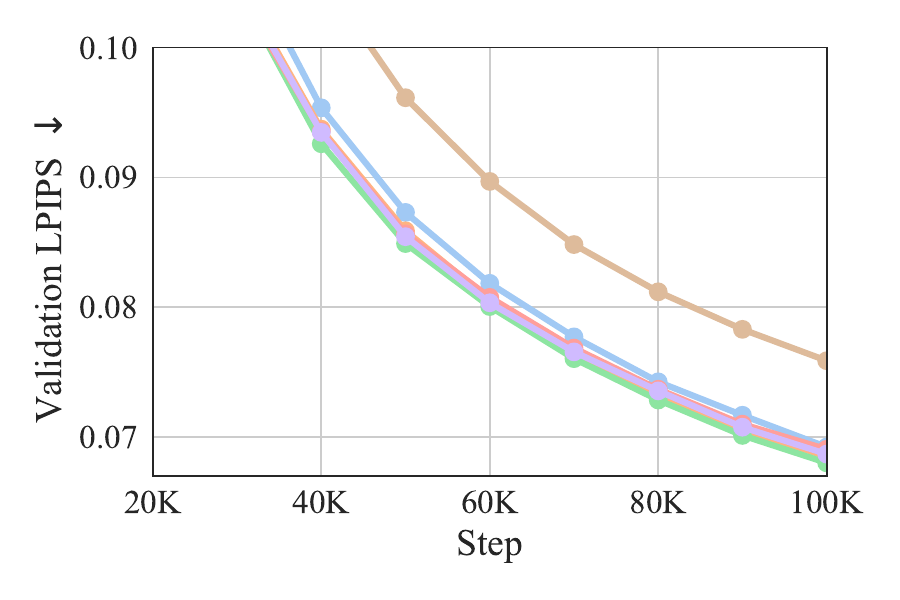}
        \vspace{-1.8em}
        \caption{WL Loss weights $\lambda_{W\!L}$.}
        \label{fig:analysis-b}
    \end{subfigure}
    \hfill
    \begin{subfigure}{0.32\textwidth}
        \centering
        \includegraphics[width=\textwidth]{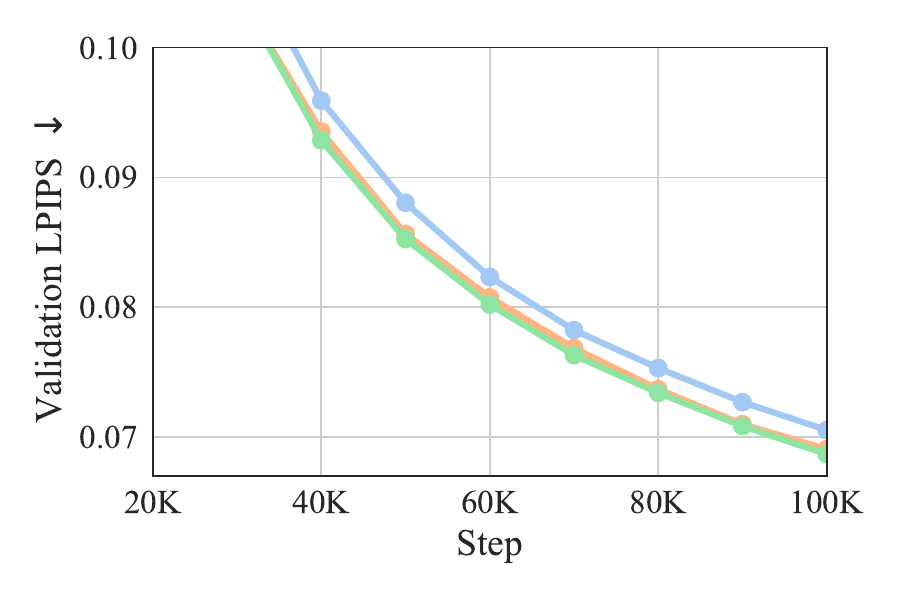}
        \vspace{-1.8em}
        \caption{Number of energy flow path channels $C_{flow}$.}
        \label{fig:analysis-c}
    \end{subfigure}
    \vspace{-0.5em}
    \caption{\textbf{Training dynamics under different settings.}}
    \label{fig:Analysis}
    \vspace{-1.2em}
\end{figure*}

\noindent\textbf{Video Reconstruction Performance.}\quad
We present a quantitative comparison of reconstruction performance between WF-VAE and baseline models in~\cref{tb:MainTable} and qualitative reconstruction results in~\cref{fig:Recon}. Despite having the lowest computational cost, WF-VAE-S outperforms popular open-source video VAEs such as OD-VAE~\cite{chen2024od} and Open-Sora VAE~\cite{opensora} on both datasets. When increasing the model complexity, WF-VAE-L competes well with Allegro~\cite{zhou2024allegroopenblackbox}, outperforming it in PSNR, LPIPS and FVD but slightly lagging in SSIM. However, WF-VAE-L is significantly more computationally efficient than Allegro. Additionally, we compare WF-VAE-L with CogVideoX~\cite{yang2024cogvideox} using 16 latent channels. Except for a slightly lower SSIM on the Webvid-10M, WF-VAE-L outperforms CogVideoX across all other metrics. These results indicate that WF-VAE achieves competitive reconstruction performance compared to state-of-the-art open-source video VAEs, while substantially improving computational efficiency.

\noindent\textbf{Video Generation Evaluation.}\quad
\cref{fig:LatteGenerate} and~\cref{tb:FVDResults} qualitatively and quantitatively demonstrate the video generation results of the diffusion model using WF-VAE. WF-VAE achieves the best performance in terms of FVD and IS metrics. For instance, in the SkyTimelapse dataset, among models with 4 latent channels, WF-VAE-S achieves the best FVD score, 10.23 lower than WF-VAE-L and 27.35 lower than OD-VAE. For models with 16 latent channels, WF-VAE-L's FVD score is 0.51 lower than CogVideoX's. This might be because the higher dimensionality of the latent space makes convergence more difficult, resulting in slightly inferior performance compared to WF-VAE with 4 latent channels under the same training steps.

\subsection{Ablation Study}\label{section:model_analysis}

\noindent\textbf{Increasing the latent dimension.}\quad
Recent works~\cite{esser2024scalingrectifiedflowtransformers,Dai_Hou_Ma_Tsai_Wang_Wang_Zhang_Vandenhende_Wang_Dubey_et} show that the number of latent channels significantly impacts reconstruction quality. We experiment with 4, 8, 16, and 32 latent channels. \cref{fig:analysis-a} illustrates that reconstruction performance improves significantly as the number of latent channels increases. However, larger latent channels may increase convergence difficulty in training diffusion model, as evidenced by the results presented in ~\cref{tb:FVDResults}.

\noindent\textbf{Exploration of WL loss weight $\lambda_{W\!L}$.}\quad
To ensure structural symmetry in WF-VAE, we introduce WL loss. As shown in \cref{fig:analysis-b}, the model substantially decreases PSNR performance when $\lambda_{W\!L}=0$.  Our experiments demonstrate optimal results for both PSNR and LPIPS metrics when $\lambda_{W\!L}=0.1$. Furthermore, our analysis of different loss functions reveals that utilizing L1 loss produces superior results compared to L2 loss.

\noindent\textbf{Expanding the energy flow path.}\quad
The low-frequency information flows into the latent representation through the energy flow pathway, and the number of channels in this path, $C_{flow}$, determines the intensity of low-frequency information injection into the backbone. We conducted experiments with $C_{flow}$ values of 64, 128, and 256. As shown in \cref{fig:analysis-c}, we find that when $C_{flow}$ is 128, it can balance reconstruction and computational performance.

\noindent\textbf{Increasing the number of base channels.}\quad
To further exploit the capabilities of the WF-VAE architecture, we increase the model complexity by expanding the number of base channels. The channel dimensionality increases by one base channel width after each downsampling layer, starting from the number of base channels. As shown in \cref{tb:comparison}, we experiment with three configurations: 128, 160, and 192 base channels. The results demonstrate that the model's performance improves as the number of base channels increases. Despite the corresponding rise in model parameters, the computational cost remains comparatively low compared to benchmark models, as shown in \cref{fig:SpeedMemoryTest}.

\begin{table}[t]
\centering
\renewcommand{\arraystretch}{1.2}
\scalebox{0.97}{
\begin{tabular}{c|ccccc}
\toprule[1.2pt]
\multirow{2}{*}{\textbf{Model}} & \multirow{2}{*}{\textbf{BC}}  & \multicolumn{2}{c}{\textbf{Params (M)}} & \multicolumn{2}{c}{\textbf{Kinetics-400}} \\
\cmidrule(lr){3-4}\cmidrule(lr){5-6}
 & & \textbf{Enc} & \textbf{Dec} & \textbf{PSNR}$\uparrow$ & \textbf{LPIPS}$\downarrow$ \\
\midrule
WF-VAE-S & 128 & 38 & 108 & 28.21 & 0.0779 \\
WF-VAE-M & 160 & 58 & 164 & \underline{28.44} & \underline{0.0699} \\
WF-VAE-L & 192 & 84 & 232 & \textbf{28.66} & \textbf{0.0661} \\
\bottomrule[1.2pt]
\end{tabular}
}
\vspace{-0.4em}
\caption{\textbf{
Scalability of WF-VAE.} We evaluated PSNR and LPIPS on Kinetics-400~\cite{kay2017kinetics}. Reconstruction performance improves as model complexity increases.}
\label{tb:comparison}
\vspace{-1.2em}
\end{table}

\noindent\textbf{Ablation studies of model architecture.}\quad
First, We evaluate the effectiveness of the energy pathway by examining its impact when removed from layer 3 alone and both layers 2 and 3. Specifically, we eliminate the wavelet transform layer and inflow-outflow module, setting the \texttt{concat} input to a zero tensor. This analysis highlights the advantages of integrating low-frequency video energy into the backbone. 
Second, we investigate the importance of the proposed WL Loss in regularizing the encoder-decoder.
Third, we analyze the effect of replacing the normalization method with layer normalization~\cite{ba2016layernormalization} for \textit{Causal Cache}.
The results of these ablation studies are shown in \cref{tb:ablation}.

\begin{table}[h!]
\centering
\renewcommand{\arraystretch}{1.10}
\scalebox{0.97}{
\begin{tabular}{c c c c c c c}
\toprule[1.2pt]
\multicolumn{5}{c}{\textbf{Settings}} & \multicolumn{2}{c}{\textbf{Kinetics-400}} \\
\cmidrule(lr){1-5}\cmidrule(lr){6-7}
\textbf{L1} & \textbf{L2} & \textbf{L3} & \textbf{WL Loss} & \textbf{NM} & \textbf{PSNR}$\uparrow$ & \textbf{LPIPS}$\downarrow$ \\
\midrule
\ding{51} & & & & L & 27.85 & 0.0737 \\
\ding{51} & \ding{51} & & \ding{51} & L & 27.94  & 0.0737 \\
\ding{51} & \ding{51} & \ding{51} &  & L & 27.90 & 0.0692 \\
\ding{51} & \ding{51} & \ding{51} & \ding{51} & L & \textbf{28.21}  & \underline{0.0690} \\
\ding{51} & \ding{51} & \ding{51} & \ding{51} & G & \underline{28.03} & \textbf{0.0684} \\
\bottomrule[1.2pt]
\end{tabular}}
\vspace{-0.2em}
 \caption{\textbf{Ablation studies on model architecture.} We evaluate the impact of three key components: energy flow pathways across network layers, WL loss, and normalization methods (L: layer normalization~\cite{ba2016layernormalization}, G: group normalization~\cite{wu2018groupnormalization}).}
 \label{tb:ablation}
 \vspace{-1.2em}
\end{table}

\subsection{Causal Cache}\label{section:casual cache}

To validate the lossless inference capability of \textit{Causal Cache}, we compare our approach with existing block-wise inference methods implemented in several open-source LVDMs. OD-VAE~\cite{chen2024od} and Allegro~\cite{zhou2024allegroopenblackbox} offer spatio-temporal tiling inference implementations, while CogVideoX~\cite{yang2024cogvideox} adopts a temporal caching strategy. As demonstrated in \cref{tb:tiling_degrade}, both tiling strategies and conventional caching methods exhibited performance degradation, while \textit{Causal Cache} achieves lossless inference with performance metrics identical to direct inference.

\begin{table}[h]
\centering
        \vspace{0.6em}
        \renewcommand{\arraystretch}{1.1}
        \scalebox{0.80}{
        \begin{tabular}{l|cccc}
            \toprule[1.2pt]
            \multirow{2}{*}{\textbf{Method}} & \multirow{2}{*}{\textbf{Chn}} & \multirow{2}{*}{\textbf{BWI}} & \multicolumn{2}{c}{\textbf{Panda70M}} \\
             \cmidrule(lr){4-5}
             & &  & \textbf{PSNR}$\uparrow$  & \textbf{LPIPS}$\downarrow$ \\
            \midrule
              & & \ding{55} & 31.71 & 0.0422 \\
              \multirow{-2}{*}{Allegro~\cite{zhou2024allegroopenblackbox}} & \multirow{-2}{*}{4} & \ding{51} & 25.31{\color{my_red} (-6.40)} & 0.1124{\color{my_red} (+0.0702)} \\
              &  & \ding{55} & 30.31 & 0.0439 \\
              \multirow{-2}{*}{OD-VAE~\cite{chen2024od}} & \multirow{-2}{*}{4} & \ding{51} & 28.51{\color{my_red} (-1.80)} & 0.0552{\color{my_red} (+0.0113)} \\
              \rowcolor{cyan!3}
              &  & \ding{55} & 32.10 & 0.0411 \\
              \rowcolor{cyan!3}
              \multirow{-2}{*}{WF-VAE-L (Ours)} & \multirow{-2}{*}{4} & \ding{51} & 32.10{\color{my_green} (0.00)} &  0.0411{\color{my_green} (0.0000)} \\
            \midrule
             \multirow{2}{*}{CogVideoX~\cite{yang2024cogvideox}} & \multirow{2}{*}{16} & \ding{55} & 35.79 & 0.0198 \\
            &  & \ding{51} & 35.41{\color{my_red} (-0.38)} & 0.0218{\color{my_red} (+0.0020)} \\
            \rowcolor{cyan!3}
            &  & \ding{55} & 35.87 & 0.0175 \\
            \rowcolor{cyan!3}
            \multirow{-2}{*}{WF-VAE-L (Ours)}  & \multirow{-2}{*}{16} & \ding{51} & 35.87{\color{my_green} (0.00)} & 0.0175{\color{my_green} (0.0000)} \\
            \bottomrule[1.2pt]
        \end{tabular}
        }
        \vspace{-0.2em}
        \captionof{table}{\textbf{Quantitative analysis of visual quality degradation induced by block-wise inference.} Values in {\color{my_red} red} indicate degradation compared to direct inference, while values in {\color{my_green} green} demonstrate preservation of quality. BWI denotes Block-Wise Inference. Experiments are conducted on 33 frames with 256×256 resolution.}
        \label{tb:tiling_degrade}
        \vspace{-1.2em}
\end{table}
\section{Conclusion}

In this paper, we propose \textbf{WF-VAE}, an innovative autoencoder that utilizes multi-level wavelet transform to extract pyramidal features, thus creating a primary energy flow pathway for encoding low-frequency video information into a latent representation. Additionally, we introduce a lossless block-wise inference mechanism called \textit{Causal Cache}, which completely resolves video flickering associated with prior tiling strategies. Our experiments demonstrate that WF-VAE achieves state-of-the-art reconstruction performance while maintaining low computational costs. WF-VAE significantly reduces the expenses associated with large-scale video pre-training, potentially inspiring future designs of video VAEs.

\noindent\textbf{Limitations and Future Work.}\quad The initial design of the decoder incorporated insights from \cite{rombach2022high}, employing a highly complex structure that resulted in more parameters
in the backbone of the decoder compared to the encoder.
Although the computational cost remains manageable, we
consider these parameters redundant. Consequently, we aim
to streamline the model in future work to fully leverage the
advantages of our architecture.

\section*{Acknowledgments}

We thank all the anonymous reviewers for their constructive comments. This work was supported in part by the Natural Science Foundation of China (No. 62202014, 62332002, 62425101, 62088102).
    
\newpage

{
    \small
    \bibliographystyle{ieeenat_fullname}
    \bibliography{main}
}


\end{document}


\clearpage
\setcounter{page}{1}

\maketitlesupplementary


The supplementary materials include further details as follows:

\begin{itemize}
    \item We present additional notations in \cref{sec:notation}.
    \item We analyze subband energy and entropy of wavelet transform in \cref{sec:wavelet}, which further validates our motivation.
    \item We present our training parameters in \cref{sec:training_details}
    \item We present the derivation of the Causal Cache formulation in \cref{sec:derivation}.
    \item We provide additional experimental results in \cref{sec:exp}.
\end{itemize}

\section{Notations}
\label{sec:notation}

The notations and their descriptions in the paper are shown in \cref{tb:notations}.

\begin{table}[h]
\centering
\renewcommand{\arraystretch}{1.0}
\scalebox{0.78}{
\begin{tabular}{>{\centering\arraybackslash}p{0.2\columnwidth}|>{\centering\arraybackslash}p{0.9\columnwidth}}
\toprule[1.2pt]
\multirow{1}{*}{\textbf{Notations}} & \multirow{1}{*}{\textbf{Descriptions}} \\
\midrule
$WT(\cdot)$ &  Wavelet transform\\
$IWT(\cdot)$ & Inverse wavelet transform\\
$\mathbf{S}^{(l)}_{\Box\Box\Box}$ & Wavelet subband within layer $l$, where $\Box\Box\Box$ specifies the type of filtering (high or low pass) applied in three dimensions. \\
$\mathcal{W}^{(l)}$ & The set of all subbands within layer $l$ \\
\bottomrule[1.2pt]
\end{tabular}
}
\caption{Notations symbols and their descriptions.}
\label{tb:notations}
\end{table}

\section{Wavelet Subband Analysis}
\label{sec:wavelet}
\begin{figure}[h]
    \centering
    \begin{subfigure}[b]{\linewidth}
        \centering
        \includegraphics[width=\linewidth]{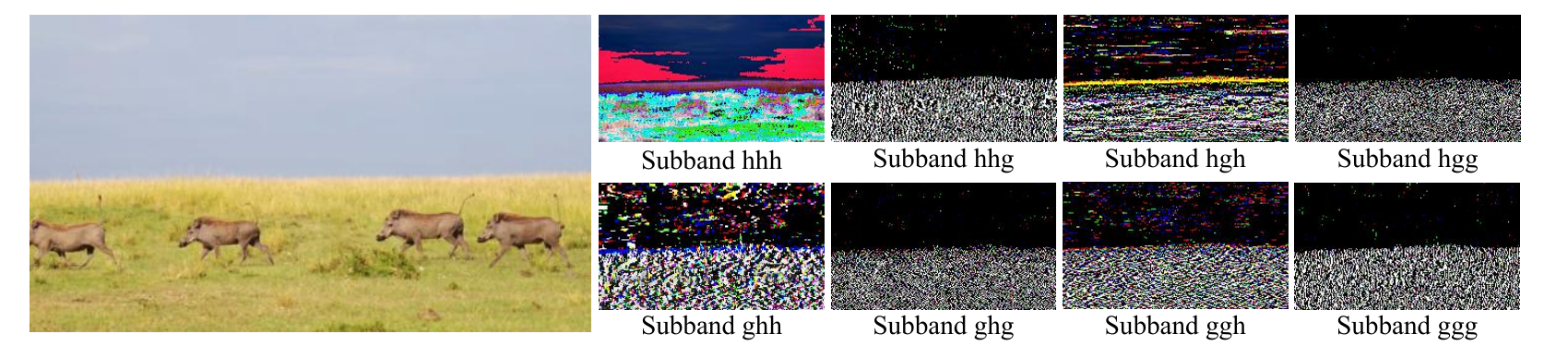}
        \caption{Visualization of the eight subbands obtained after wavelet transform of the video.}
    \end{subfigure}
    
    \begin{subfigure}[b]{\linewidth}
        \centering
        \includegraphics[width=\linewidth]{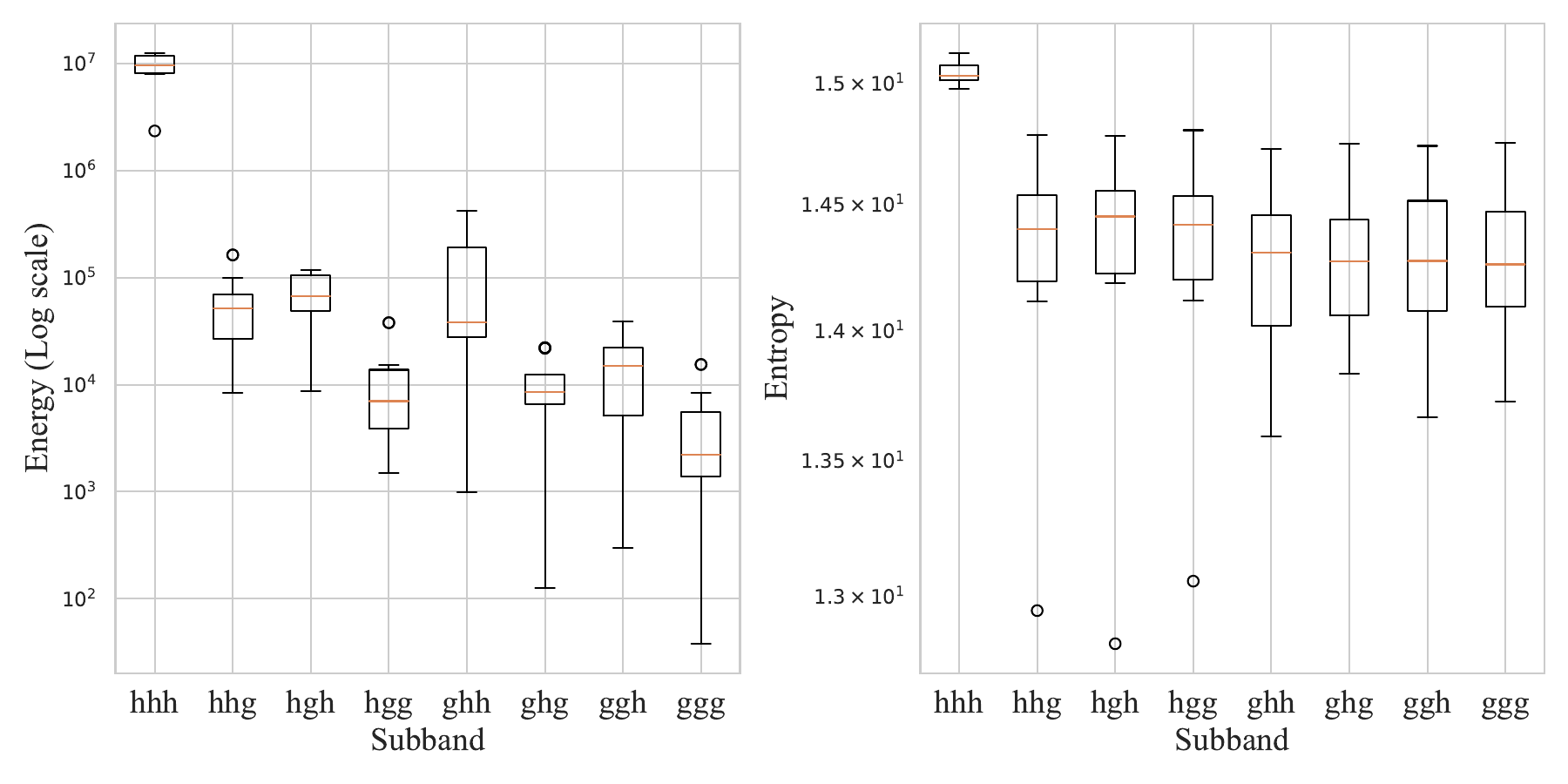}
        \caption{Energy and entropy of each subband.}
        \label{fig:ana-b}
    \end{subfigure}
    \caption{Visualization of the subbands and their respective energy and entropy.}
\end{figure}

We analyze the energy and entropy distributions across the subbands obtained after wavelet transform. As illustrated in \cref{fig:ana-b}, the energy and entropy of the video are primarily concentrated in the $hhh$ low-frequency subband. This concentration suggests that low-frequency components carry more significant information and necessitate lower compression rates to ensure superior reconstruction performance. This observation further validates the rationale behind our proposed approach.

\section{Training Details}
\label{sec:training_details}

The training hyperaparameters are shown in \cref{tb:train_hyper}.

\begin{table}[h]
\centering
\renewcommand{\arraystretch}{1.0}
\begin{tabular}{>{\centering\arraybackslash}p{0.3\textwidth}c}  
\toprule[1.2pt]
\multirow{1}{*}{\textbf{Parameter}} & \multirow{1}{*}{\textbf{Setting}} \\
\midrule
\multicolumn{2}{c}{\textit{Stage I - 800k step}} \\
\midrule
Learning Rate & 1e-5 \\
Total Batch Size & 8 \\
Peceptual(LPIPS) Weight  & 1.0 \\
WL Loss Weight ($\lambda_{W\!L} $) & 0.1 \\
KL Weight ($\lambda_{K\!L} $) & 1e-6 \\
Resolution & 256$\times$256 \\
Num Frames & 25 \\
EMA Decay & 0.999 \\
\midrule
\multicolumn{2}{c}{\textit{Stage II - 200k step}} \\
\midrule
Num Frames & 49 \\
\midrule
\multicolumn{2}{c}{\textit{Stage III - 200k step}} \\
\midrule
Peceptual(LPIPS) Weight  & 0.1 \\
\bottomrule[1.2pt]
\end{tabular}
\caption{Training hyperparameters across three stages.}
\label{tb:train_hyper}
\end{table}

\begin{figure*}[ht]
    \centering    \includegraphics[width=0.95\linewidth]{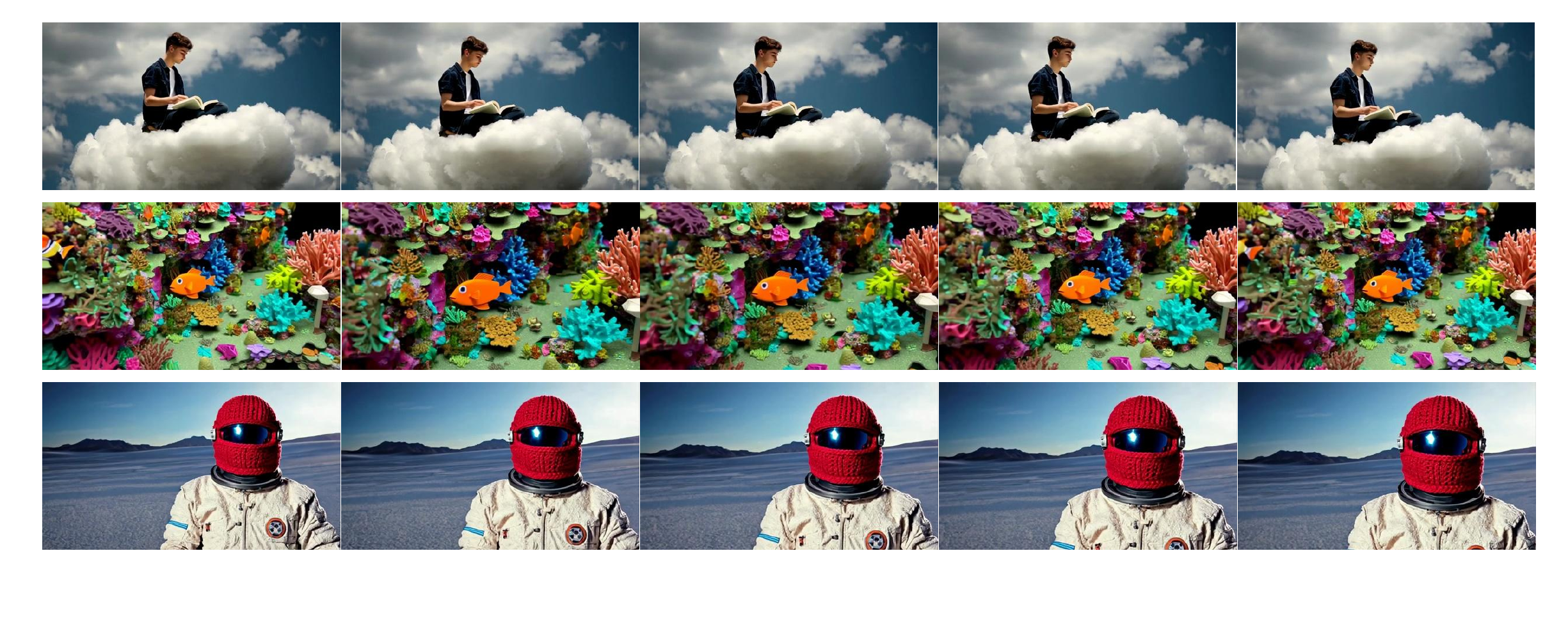}
    \vspace{-2.5em}
    \caption{\textbf{Qualitative experiments in video generation model pretraining.} It demonstrates that WF-VAE can be effectively applied to the training of downstream diffusion models.}
    \label{fig:pretrain}
\end{figure*}

 \begin{table*}[h]
    \centering
    \scalebox{1.0}{
    \begin{tabular}{lcccccccc}
        \toprule
        Method & \multicolumn{4}{c}{480P} & \multicolumn{4}{c}{720P} \\
        \cmidrule(lr){2-5} \cmidrule(lr){6-9}
         & PSNR$\uparrow$ & LPIPS$\downarrow$ & rFVD$\downarrow$ & SSIM$\uparrow$ & PSNR$\uparrow$ & LPIPS$\downarrow$ & rFVD$\downarrow$ & SSIM$\uparrow$\\
        \midrule
        \multicolumn{9}{c}{\textit{4 latent channels}} \\
        \midrule
        \rowcolor{cyan!3}
        WF-VAE-L & \textbf{30.56} & \textbf{0.0595} & \textbf{55.65} & \textbf{0.8713} & \textbf{31.12} & \textbf{0.0617} & \textbf{49.93} & \textbf{0.8799} \\
        Allegro & 30.06 & 0.0689 & 105.70 & 0.8673 & 30.78 & 0.0668 & 86.85 & 0.8795 \\
        \midrule
        \multicolumn{9}{c}{\textit{16 latent channels}} \\
        \midrule
        \rowcolor{cyan!3}
        WF-VAE-L & \textbf{34.28} & \textbf{0.0275} & \textbf{20.43} & \textbf{0.9347} & \textbf{34.82} & \textbf{0.0294} & \textbf{19.27} & \textbf{0.9384} \\
        CogVideoX & 33.85 & 0.0317 & 32.85 & 0.9319 & 34.24 & 0.0331 & 24.82 & 0.9364 \\
        \bottomrule
    \end{tabular}}
    \caption{Quantitative evaluation on Inter4K dataset, using \textbf{65 frames}.}
    \label{table:exp1}
\end{table*}

\section{Derivation of Causal Cache}
\label{sec:derivation}

\begin{figure}[h]
    \centering    \includegraphics[width=\linewidth]{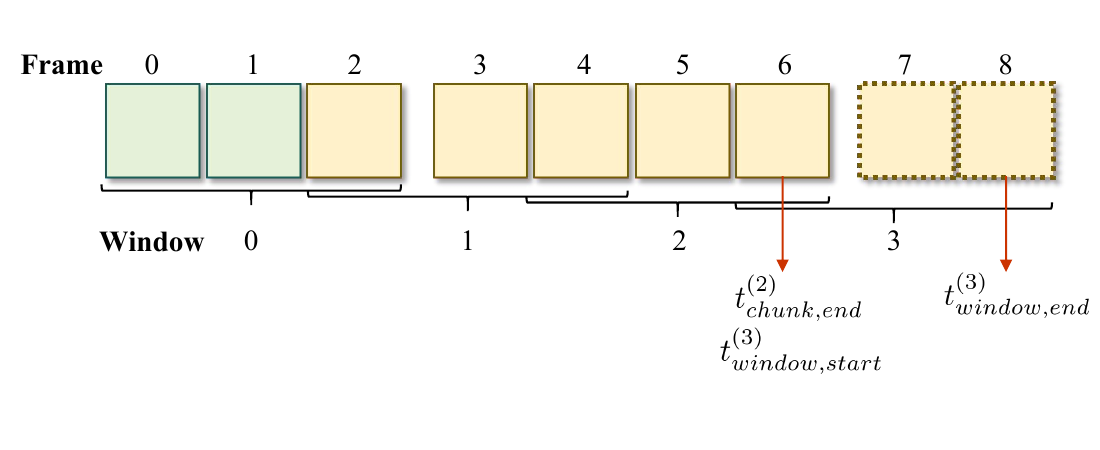}
    \caption{Illustration of \textit{Causal Cache} with parameters $k$=3, $s$=2, and chunk size $T_{chunk}$=4.}
    \label{fig:causalcache}
\end{figure}

Let us define a convolution with sliding window index $n \in \mathbb{N}_0$ and chunk index $m \in \mathbb{N}_0$. Given a convolutional stride $s$ and kernel size $k$, as shown in \cref{fig:causalcache}, the starting and ending frame indices for each sliding window are:
\begin{align}
t_{window,start}(n) &= ns, \\
t_{window,end}(n) &= t_{window,start}(n) + k - 1.
\end{align}

For chunk boundaries, we define:
\begin{equation}
t_{chunk,end}(m) = k - 1 + mT_{chunk}
\end{equation}

where $T_{chunk}$ denotes the chunk size. For a given chunk index $m$, the maximum sliding index $n_{max}(m)$ is determined by the constraint $t_{window,end}(n) \geq t_{chunk,end}(m)$:

\begin{equation}
n_{\text{max}}(m) = \left\lfloor\frac{mT_{chunk}}{s} + 1\right\rfloor.
\end{equation}

Consequently, the required cache size $T_{\text{cache}}(m)$ for chunk $m$ is:

\begin{equation}
\begin{split}
T_{\text{cache}}(m) = t_{\text{chunk,end}}(m) - t_{\text{window,start}}(n_{\text{max}}(m))\\  + 1
= mT_{\text{chunk}} + k - \left\lfloor\frac{mT_{\text{chunk}}}{s} + 1\right\rfloor s    
\end{split}
\end{equation}

\section{Additional Experiments}
\label{sec:exp}

\noindent\textbf{Evaluation Across Different Resolutions.}\quad  To validate the robustness of WF-VAE across different resolutions, we conduct metric evaluations on the Inter4K dataset at 480P and 720P resolutions. As shown in \cref{table:exp1}, WF-VAE demonstrates competitive performance in reconstruction tasks across varying resolutions.

\noindent\textbf{Validation in Diffusion Model Pretraining.} To verify the applicability of WF-VAE to LVDM, we select Open-Sora Plan for pretraining on large-scale datasets with a resolution of 512$\times$288 pixels. Qualitative experiments, as illustrated in \ref{fig:pretrain}, demonstrate promising generative performance.

\begin{figure*}[ht]
    \centering    \includegraphics[width=0.95\linewidth]{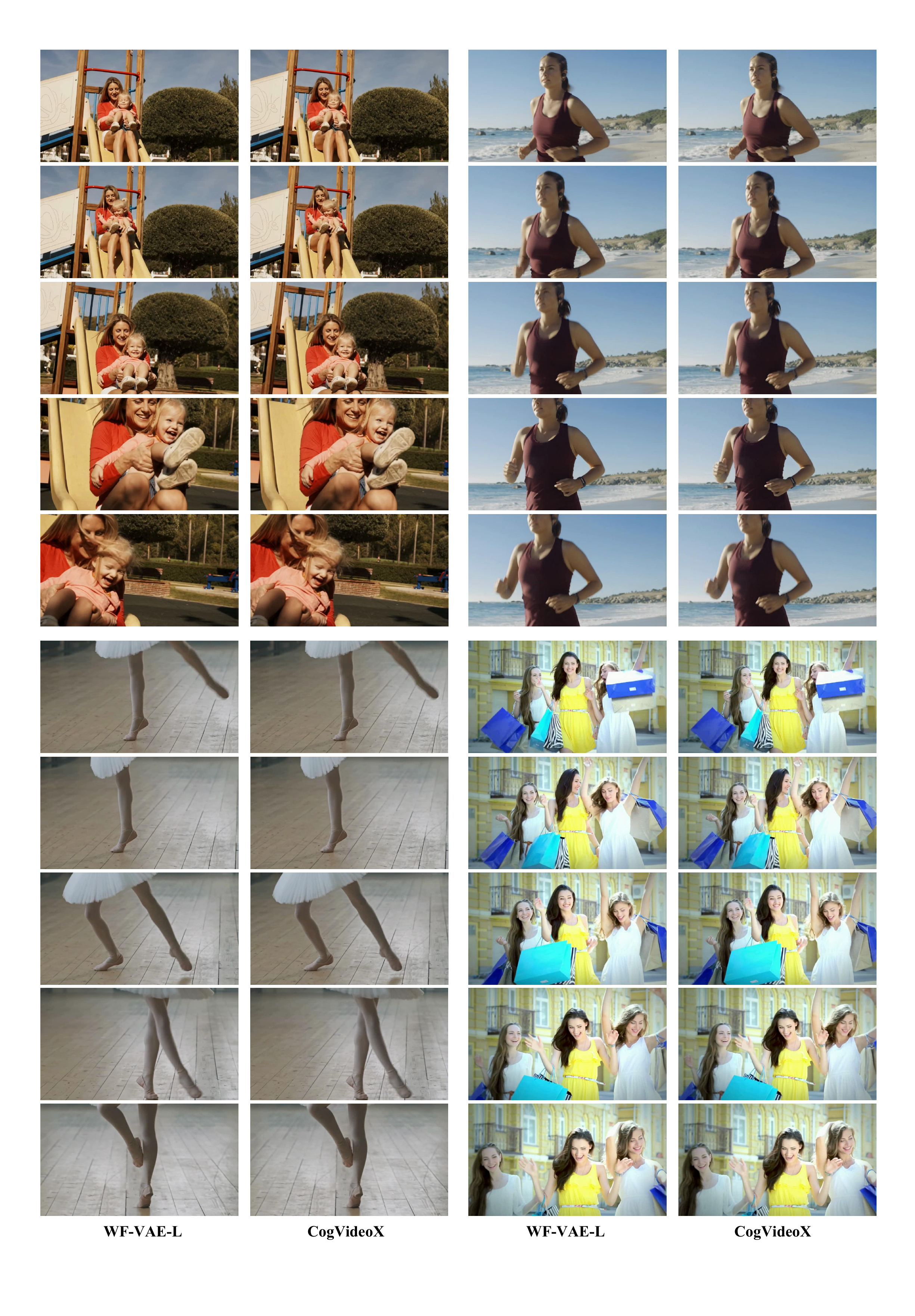}
    \vspace{-3.5em}
    \caption{\textbf{Qualitative experiments in high motion videos.} We include more 480P comparison videos in the supplementary materials.}
    \label{fig:videos}
\end{figure*}

\noindent\textbf{More Qualitative Evaluations.} To further demonstrate the capability of our model in achieving state-of-the-art reconstruction performance with low computational cost, we conduct additional qualitative evaluations against the representative VAE, CogVideoX. Refer to \cref{fig:videos} and supplementary material for more video examples.

